\documentclass[runningheads]{llncs}

\usepackage{booktabs}

\usepackage[T1]{fontenc}
\usepackage{amsmath}
\usepackage{amsfonts}
\usepackage{graphicx}
\usepackage[table,xcdraw]{xcolor}
\usepackage{multirow}
\usepackage[pagebackref=true,breaklinks=true,letterpaper=true,colorlinks,bookmarks=false]{hyperref}
\usepackage{bbm}
\usepackage{color}

\usepackage{caption}
\usepackage{subcaption}

\begin{document}
\title{Source-Free Domain Adaptive Fundus Image Segmentation with Class-Balanced Mean Teacher}

\titlerunning{Class-Balanced Mean Teacher}
% If the paper title is too long for the running head, you can set
% an abbreviated paper title here
%

% \author{Anonymous}
% \institute{Anonymous Organization \\\textbf{}
% \email{**@******.***}
% }

\author{Longxiang Tang\inst{1} \and
%index{Tang, Longxiang}
Kai Li\inst{2}\protect\footnotemark[1] \and
%index{Li, Kai}
Chunming He\inst{1} \and
%index{He, Chunming}
Yulun Zhang\inst{3} \and
%index{Zhang, Yulun}
Xiu Li\inst{1}\protect\footnotemark[1]}
%index{Li, Xiu}

\authorrunning{L. Tang et al.}
% First names are abbreviated in the running head.
% If there are more than two authors, 'et al.' is used.
%
\institute{Tsinghua Shenzhen International Graduate School, Tsinghua University, China \\
% \email{li.xiu@sz.tsinghua.edu.cn} 
\and
NEC Laboratories America, USA
% \email{kaili@nec-labs.com}
\and
ETH Zurich, Switzerland \\
\email{\{lloong.x, li.gml.kai, chunminghe19990224, yulun100\}@gmail.com li.xiu@sz.tsinghua.edu.cn}
} 
\maketitle              % typeset the header of the contribution

\renewcommand{\thefootnote}{\fnsymbol{footnote}} %将脚注符号设置为fnsymbol类型，即特殊符号表示
\footnotetext[1]{Corresponding author.} %对应脚注[2]
\renewcommand{\thefootnote}{\arabic{footnote}}

%% Despite 

\begin{abstract}
This paper studies source-free domain adaptive fundus image segmentation which aims to adapt a pretrained fundus segmentation model to a target domain using unlabeled images. This is a challenging task because it is highly risky to adapt a model only using unlabeled data. Most existing methods tackle this task mainly by designing techniques to carefully generate pseudo labels from the model's predictions and use the pseudo labels to train the model. While often obtaining positive adaption effects, these methods suffer from two major issues. First, they tend to be fairly unstable - incorrect pseudo labels abruptly emerged may cause a catastrophic impact on the model. Second, they fail to consider the severe class imbalance of fundus images where the foreground (e.g., cup) region is usually very small. This paper aims to address these two issues by proposing the Class-Balanced Mean Teacher (CBMT) model. CBMT addresses the unstable issue by proposing a weak-strong augmented mean teacher learning scheme where only the teacher model generates pseudo labels from weakly augmented images to train a student model that takes strongly augmented images as input. The teacher is updated as the moving average of the instantly trained student, which could be noisy. This prevents the teacher model from being abruptly impacted by incorrect pseudo-labels. For the class imbalance issue, CBMT proposes a novel loss calibration approach to highlight foreground classes according to global statistics. Experiments show that CBMT well addresses these two issues and outperforms existing methods on multiple benchmarks.

\end{abstract}

\setlength{\abovedisplayskip}{2pt}
\setlength{\belowdisplayskip}{2pt}

\section{Introduction}
Medical image segmentation plays an essential role in computer-aided diagnosis systems in different applications and has been tremendously advanced in the past few years~\cite{ronneberger2015u,milletari2016v,drozdzal2016importance,ibtehaz2020multiresunet}. While the segmentation model~\cite{ren2015faster,carion2020end,he2023camouflaged,he2023weakly} always requires sufficient labeled data, unsupervised domain adaptation (UDA) approaches have been proposed, learning an adaptive model jointly with unlabeled target domain images and labeled source domain images \cite{ganin2016domain}, for example, the adversarial training paradigm\cite{kamnitsas2017unsupervised,javanmardi2018domain,cai2019towards,gadermayr2019generative,li2020adversarial}.

Although impressive performance has been achieved, these UDA methods may be limited for some real-world medical image segmentation tasks where labeled source images are not available for adaptation. This is not a rare scenario because medical images are usually highly sensitive in privacy and copyright protection such that  
% annotating them usually demands expert knowledge. For either privacy or copyright protection, 
labeled source images may not be allowed to be distributed. This motivates the investigation of source-free domain adaptation (SFDA) where adapts a source segmentation model trained on labeled source data (in a private-protected way) to the target domain only using unlabeled data.

% a new learning paradigm, adapting a pre-trained model to a new domain with a unlabeled data without source data, which is also called \emph{source-free domain adaptation}. This task is more challenging for its lack of direct access to source data distribution, but on the other hand, it's more practical.

A few recent SFDA works have been proposed. 
% perform domain adaptation in the absence of source data. 
OSUDA~\cite{liu2021adapting} utilizes the domain-specific low-order batch statistics and domain-shareable high-order batch statistics, trying to adapt the former and keep the consistency of the latter. SRDA~\cite{bateson2020source} minimizes a label-free entropy loss guided with a domain-invariant class-ratio prior.
DPL~\cite{chen2021source} introduces pixel-level and class-level pseudo-label denoising schemes to reduce noisy pseudo-labels and select reliable ones. 
U-D4R~\cite{xu2022denoising} applies an adaptive class-dependent threshold with the uncertainty-rectified correction to realize better denoising.

\begin{figure}[t]
\includegraphics[width=\textwidth]{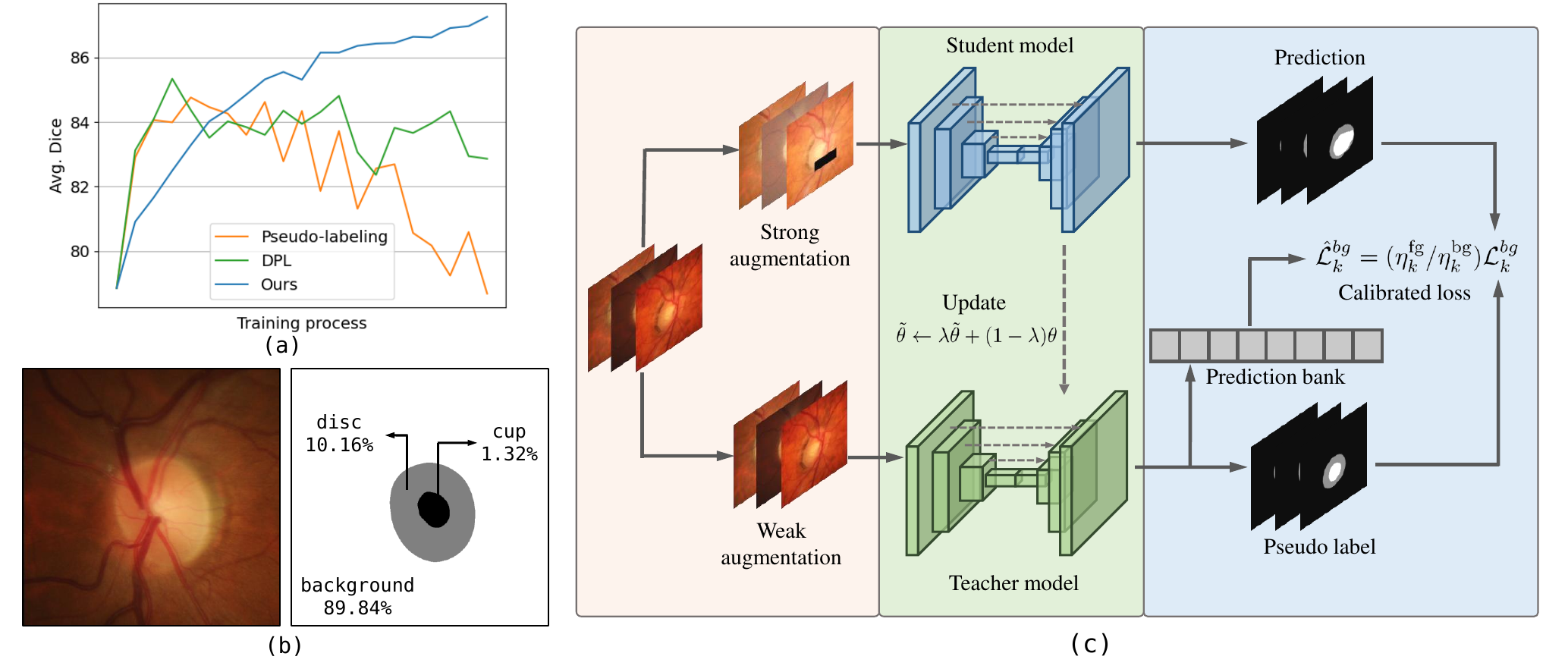}
% \vspace{-0.7cm}
\caption{(a) Training curve of vanilla pseudo-labeling, DPL~\cite{chen2021source} and our approach. (b) Fundus image and its label with class proportion from the \textit{RIM-ONE-r3} dataset. (c) Illustrated framework of our proposed CBMT method.} 
\label{fig1}
% \vspace{-0.4cm}
\end{figure}

Although these methods have achieved some success in model adaptation, they still suffer from two major issues. \underline{First}, they tend to be fairly unstable. Without any supervision signal from labeled data, the model heavily relies on the predictions generated by itself, which are always noisy and could easily make the training process unstable, causing catastrophic error accumulation after several training epochs as shown in Fig.~\ref{fig1}(a).  Some works avoid this problem by only training the model for very limited iterations (only 2 epochs in \cite{chen2021source,xu2022denoising}) and selecting the best-performing model during the whole training process for testing. However, this does not fully utilize the data and it is non-trivial to select the best-performing model for this unsupervised learning task. \underline{Second}, they failed to  consider the severe foreground and background imbalance of fundus images where the foreground (e.g., cup) region is usually very small (as shown in Fig.~\ref{fig1}(b)). This oversight could also lead to a model degradation due to the dominate background learning signal.

In this paper, we propose the Class-Balanced Mean Teacher (CBMT) method to address the limitations of existing methods. To mitigate the negative impacts of incorrect pseudo labels, we propose a weak-strong augmented mean teacher learning scheme which involves a teacher model and a student model that are both initialized from the source model. We use the teacher to generate pseudo label from a weakly augmented image, and train the student that takes strongly augmented version of the same image as input. We do not train the teacher model directly by back-propagation but update its weights as the moving average of the student model. This prevents the teacher model from being abruptly impacted by incorrect pseudo labels and meanwhile accumulates new knowledge learned by the student model. 
To address the imbalance between foreground and background, 
we propose to calibrate the segmentation loss and highlight the foreground class, based on the prediction statistics derived from the global information. We maintain a prediction bank to capture global information, which is considered more reliable than that inside one image.

Our contributions can be summarized as follows: \underline{(1)} We propose the weak-strong augmented mean teacher learning scheme to address the stable issue of existing methods. \underline{(2)} We propose the novel global knowledge-guided loss calibration technique to address the foreground and background imbalance problem. \underline{(3)} Our proposed CBMT reaches state-of-the-art performance on two popular benchmarks for adaptive fundus image segmentation.

\section{Method}
Source-Free Domain Adaptive (SFDA) fundus image segmentation aims to adapt a source model $h$, trained with $N_S$ labeled source images $\mathcal{S}=\{(X_i, Y_i)\}_{i=1}^{N_S}$, to the target domain using only $N_T$ unlabeled target images $\mathcal{T}=\{X_i\}_{i=1}^{N_T}$. $Y_i\in \{0, 1\}^{H\times W \times C}$ is the ground truth, and $H$, $W$, and $C$ denote the image height, width, and class number, respectively.   
% indicating which category a pixel belongs to from ``disc'' ``cup'' and ``background''.
A vanilla pseudo-labeling-based method generates pseudo labels
$\hat{y}\in \mathbb{R}^C$ from the sigmoided model prediction $p=h(x)$ for each pixel $x\in X_i$ with source model $h$:
\begin{equation}
\label{eq1}
\hat{y}_{k}=\mathbbm{1}\left[p_k > \gamma\right],
\end{equation}
where $\mathbbm{1}$ is the indicator function and $\gamma\in[0, 1]$ is the probability threshold for transferring soft probability to hard label. $p_k$ and $y_k$ is the $k$-th dimension of $p$ and $y$, respectively, denoting the prediction and pseudo label for class $k$.
% denote the prediction and pseudo ground truth for class $k$. 
% for the $k$-th class, i.e., the $k$-th dimension of $p=h(x)$ prediction
% where , and $f_k$ is model probability score output for class $k$. 
Then $(x,\hat{y})$ is utilized to train the source model $h$ with binary cross entropy loss:
\begin{equation}
% \label{eq1}
L_{bce}= \mathbb{E}_{x\sim X_i} [\hat{y}\log(p)+(1-\hat{y})\log(1-p)]
\label{bce}
\end{equation}

Most existing SFDA works refine this vanilla method by proposing techniques to calibrate $p$ and get better pseudo label $\hat{y}$, or measure the uncertainty of $p$ and apply a weight when using $\hat{y}$ for computing the loss  \cite{chen2021source,xu2022denoising}. While achieving improved performance, these methods still suffer from the unstable issue because noisy $\hat{y}$ will directly impact $h$, and the error will accumulate since then the predictions of $h$ will be used for pseudo labeling. Another problem with this method is that they neglect the imbalance of the foreground and background pixels in fungus images, where the foreground region is small. Consequently, the second term in Eq. \eqref{bce} will dominate the loss, which is undesirable. 

% the unstable problem that  because any noises in  will directly impact $h$$
Our proposed CBMT model addresses the two problems by proposing the weak-strong augmented mean teacher learning scheme and the global knowledge-guided loss calibration technique. 
% model achieves this goal by conducting an EMA model update strategy and global knowledge class loss calibration.   
Fig. \ref{fig1}(c) shows the framework of CBMT.

% \vspace{-2mm}
\subsection{Weak-Strong Augmented Mean Teacher}
% \vspace{-1mm}
To avoid error accumulation and achieve a robust training process, we introduce the weak-strong augmented mean teacher learning scheme where there is a teacher model $h_t$ and a student model $h_s$ both initialized from the source model $h$. We generate pseudo labels with $h_t$ and use the pseudo labels to train $h_s$. To enhance generalization performance, we further introduce a weak-strong augmentation mechanism that feeds weakly and strongly augmented images to the teacher model and the student model, respectively.

Concretely, for each image $X_i$, we generate a weakly-augmented version $X^w_i$ by using image flipping and resizing. Meanwhile, we generate a 
strongly-augmented version $X^s_i$. The strong augmentations we used include a random eraser, contrast adjustment, and impulse noises.
For each pixel $x^w\in X^w_i$, we generate pseudo label $\hat{y}^w=h_t(x)$ by the teacher model $h_t$ with Eq. \eqref{eq1}. Then, we train the student model $h_s$ with 
\begin{equation}
\mathcal{L} = \mathbb{E}_{x^s\sim X^s_i, \hat{y}^w} [\tilde{\mathcal{L}}_{bce}],
\label{total_loss}
\end{equation}
where $\tilde{\mathcal{L}}_{bce}$ is the refined binary cross entropy loss which we will introduce later. It is based on Eq. \eqref{bce} but addresses the fore- and back-ground imbalance problem.  

The weakly-strong augmentation mechanism has two main benefits. First,  
since fundus image datasets are always on a small scale, the model could easily get overfitted due to the insufficient training data issue. To alleviate it, we enhance the diversity of the training set by introducing image augmentation techniques. 
Second, learning with different random augmentations performs as a consistency regularizer constraining images with similar semantics to the same class, which forms a more distinguishable feature representation.

We update the student model by back-propagating the loss defined in Eq. \eqref{total_loss}. But for the teacher model, we update it as the exponential moving average (EMA) of the student model as,
\begin{equation}
\label{eq3}
\tilde{\theta}\leftarrow \lambda\tilde{\theta}+(1-\lambda)\theta,
\end{equation}
where $\tilde{\theta}$, $\theta$ are the teacher and student model weights separately. Instead of updating the model with gradient directly, we define the teacher model as the exponential moving average of students, which makes the teacher model more consistent along the adaptation process. With this, we could train a model for a relatively long process and safely choose the final model without accuracy validation. From another perspective, the teacher model can be interpreted as a temporal ensemble of students in different time steps \cite{liu2021unbiased}, which enhances the robustness of the teacher model.

% \vspace{-2mm}
\subsection{Global Knowledge Guided Loss Calibration}
\label{loss_calibration}
For a fundas image, the foreground object (e.g., cup) is usually quite small and most pixel will the background. If we update the student model with Eq. \eqref{bce}, the background class will dominate the loss, which dilutes the supervision signals for the foreground class. The proposed global knowledge guided loss calibration technique aims to address this problem.           

A naive way to address the foreground and background imbalance is to calculate the numbers of pixels falling into the two categories, respectively, within each individual image and devise a loss weighting function based on the numbers. This strategy may work well for the standard supervised learning tasks, where the labels are reliable. But with pseudo labels, it is too risky to conduct the statistical analysis based on a single image. To remedy this, we analyze the class imbalance across the whole dataset, and use this global knowledge to calibrate our loss for each individual image.

Specifically,  we store the predictions of pixels from all images and maintain the mean loss for foreground and background as,
\begin{equation}
\label{eq4}
\eta_k^{\text{fg}} = \frac{\sum_{i}\mathcal{L}_{i,k}\cdot\mathbbm{1}[\hat{y}_{i,k}=1]}{\sum_{i}\mathbbm{1}[\hat{y}_{i,k}=1]};\ \ 
\eta_k^{\text{bg}} = \frac{\sum_{i}\mathcal{L}_{i,k}\cdot\mathbbm{1}[\hat{y}_{i,k}=0]}{\sum_{i}\mathbbm{1}[\hat{y}_{i,k}=0]}
\end{equation}
where $\mathcal{L}$ is the segmentation loss mentioned above, and ``fg'' and ``bg'' represent foreground/background. The reason we use the mean of the loss, rather than the number of pixels, is that the loss of each pixel indicates the ``hardness`` of each pixel according to the pseudo ground truth. This gives more weight to those more informative pixels, thus more global knowledge is considered.

With each average loss, the corresponding learning scheme could be further calibrated. We utilize the ratio of $\eta_k^{\text{fg}}$ to $\eta_k^{\text{bg}}$ to weight background loss $\mathcal{L}_k^{bg}$:
\begin{equation}
\tilde{\mathcal{L}}_{bce}= \mathbb{E}_{x\sim X_i, k\sim C} [\hat{y}_k\log(p_k)+\eta_k^{\text{fg}}/\eta_k^{\text{bg}}(1-\hat{y_k})\log(1-p_k)]
\label{calibrated_bce}
\end{equation}

The calibrated loss ensures fair learning among different classes, therefore alleviating model degradation issues caused by class imbalance.

Since most predictions are usually highly confident (very close to 0 or 1), they are thus less informative.  We need to only include pixels with relatively large loss scales to compute mean loss. We realize this by adopting constraint threshold $\alpha$ to select pixels: $\frac{|f(x_i)-\gamma|}{|\hat{y_i}-\gamma|}>\alpha$, where $\alpha$ is set by default to 0.2. $\alpha$ represents the lower bound threshold of normalized prediction, which can filter well-segmented uninformative pixels out.

% \vspace{-1mm}
\section{Experiments}
% \vspace{-1mm}
\noindent\textbf{Implementation details\footnotemark[1]}.
\footnotetext[1]{The code can be found in \url{https://github.com/lloongx/SFDA-CBMT}}
We apply the Deeplabv3+ \cite{chen2018encoder} with MobileNetV2 \cite{sandler2018mobilenetv2} backbone as our segmentation model, following the previous works \cite{wang2019boundary,chen2021source,xu2022denoising} for a fair comparison. For model optimization, we use Adam optimizer with 0.9 and 0.99 momentum coefficients. During the source model training stage, the initial learning rate is set to 1e-3 and decayed by 0.98 every epoch, and the training lasts 200 epochs. At the source-free domain adaptation stage, the teacher and student model are first initialized by the source model, and the EMA update scheme is applied between them for a total of 20 epochs with a learning rate of 5e-4. Loss calibration parameter $\eta$ is computed every epoch and implemented on the class cup. The output probability threshold $\gamma$ is set as 0.75 according to previous study \cite{wang2019boundary} and model EMA update rate $\lambda$ is 0.98 by default. We implement our method with PyTorch on one NVIDIA 3090 GPU and set batch size as 8 when adaptation.

\begin{table}[t]
	\centering
	\caption{Quantitative results of comparison with different methods on two datasets, and the best score for each column is highlighted. - means the results are not reported by that method, $\pm$ refers to the standard deviation across samples in the dataset. S-F means source-free.}
	% \vspace{-0.3cm}
	\label{tab1}
	\begin{center}
		\setlength\tabcolsep{3.0pt}
		\resizebox{0.9\textwidth}{!}{%
			\begin{tabular}{l|c|c|c|c|c}
				\toprule[1.5pt]
				
				\multirow{2}{*}{Methods} & \multirow{2}{*}{S-F} & \multicolumn{2}{c|}{Optic Disc Segmentation} & \multicolumn{2}{c}{Optic Cup Segmentation} \\
				\cline{3-6}
				{} & {} & Dice[\%]\ $\uparrow$ & ASSD[pixel]\ $\downarrow$ & Dice[\%]\ $\uparrow$ & ASSD[pixel]\ $\downarrow$ \\
				\cline{1-6}
				\multicolumn{6}{l}{\emph{RIM-ONE-r3}}\\
				\hline
				W/o DA~\cite{chen2021source} & & 83.18$\pm${6.46} & 24.15$\pm${15.58} & 74.51$\pm${16.40} & 14.44$\pm${11.27} \\
				Oracle~\cite{wang2019boundary} & & 96.80 & \textbf{--} & 85.60 & \textbf{--} \\
				\hline
				\hline
				BEAL~\cite{wang2019boundary} & $\times$ & 89.80 & \textbf{--} & 81.00 & \textbf{--} \\
				AdvEnt~\cite{vu2019advent} & $\times$ & 89.73$\pm${3.66} & 9.84$\pm${3.86} & 77.99$\pm${21.08} & \textbf{7.57}$\pm${4.24} \\
				SRDA~\cite{bateson2020source} & $\checkmark$ & 89.37$\pm${2.70} & 9.91$\pm${2.45} & 77.61$\pm${13.58} & 10.15$\pm${5.75} \\
				DAE~\cite{karani2021test} & $\checkmark$ & 89.08$\pm${3.32} & 11.63$\pm${6.84} & 79.01$\pm${12.82} & 10.31$\pm${8.45} \\
				DPL~\cite{chen2021source} & $\checkmark$ & 90.13$\pm${3.06} & 9.43$\pm${3.46} & 79.78$\pm${11.05} & 9.01$\pm${5.59} \\
				\hline
				\textbf{CBMT(Ours)} & $\checkmark$ & \textbf{93.36$\pm${4.07}} & \textbf{6.20$\pm${4.79}} & \textbf{81.16$\pm${14.71}} & 8.37$\pm${6.99} \\
				\hline

				\multicolumn{6}{l}{\emph{Drishti-GS}}\\
				\hline
				W/o DA~\cite{chen2021source} & & 93.84$\pm${2.91} & 9.05$\pm${7.50} & 83.36$\pm${11.95} & 11.39$\pm${6.30} \\
				Oracle~\cite{wang2019boundary} & & 97.40 & \textbf{--} & 90.10 & \textbf{--} \\
				\hline
				\hline
				BEAL~\cite{wang2019boundary} & $\times$ & 96.10 & \textbf{--} & \textbf{86.20} & \textbf{--} \\
				AdvEnt~\cite{vu2019advent} & $\times$ & 96.16$\pm${1.65} & 4.36$\pm${1.83} & 82.75$\pm${11.08} & 11.36$\pm${7.22} \\
				SRDA~\cite{bateson2020source} & $\checkmark$ & 96.22$\pm${1.30} & 4.88$\pm${3.47} & 80.67$\pm${11.78} & 13.12$\pm${6.48} \\
				DAE~\cite{karani2021test} & $\checkmark$ & 94.04$\pm${2.85} & 8.79$\pm${7.45} & 83.11$\pm${11.89} & 11.56$\pm${6.32} \\
				DPL~\cite{chen2021source} & $\checkmark$ & 96.39$\pm${1.33} & \textbf{4.08$\pm${1.49}} & 83.53$\pm${17.80} &11.39$\pm${10.18} \\
				\hline
				\textbf{CBMT(Ours)} & $\checkmark$ & \textbf{96.61$\pm$1.45} & \textbf{3.85$\pm$1.63} & 84.33$\pm$11.70 & \textbf{10.30$\pm$5.88} \\
				% \hline
				
				\bottomrule[1.5pt]
		    \end{tabular}}
	\end{center}
	% \vspace{-0.7cm}
\end{table}

\noindent\textbf{Datasets and metrics}.
% To perform a comprehensive study with existing UDA and SFDA methods, 
We evaluate our method on widely-used fundus optic disc and cup segmentation datasets from different clinical centers. Following previous works, We choose the REFUGE challenge training set \cite{orlando2020refuge} as the source domain and adapt the model to two target domains: RIM-ONE-r3 \cite{fumero2011rim} and Drishti-GS \cite{sivaswamy2015comprehensive} datasets for evaluation. Quantitatively, the source domain consists of 320/80 fundus images for training/testing with pixel-wise optic disc and cup segmentation annotation, while the target domains have 99/60 and 50/51 images. Same as \cite{wang2019boundary}, the fundus images are cropped to $512\times 512$ as ROI regions.

We compare our CBMT model with several state-of-the-art domain adaptation methods, including UDA methods BEAL~\cite{wang2019boundary} and AdvEnt~\cite{vu2019advent} and SFDA methods: SRDA~\cite{bateson2020source}, DAE~\cite{karani2021test} and DPL~\cite{chen2021source}. More comparisons with U-D4R~\cite{xu2022denoising} under other adaptation settings could be found in supplementary materials.
General metrics for segmentation tasks are used for model performance evaluation, including the Dice coefficient and Average Symmetric Surface Distance (ASSD). The dice coefficient (the higher the better) gives pixel-level overlap results, and ASSD (the lower the better) indicates prediction boundary accuracy.

\begin{table}[t]
	\centering
	\caption{Ablation study results of our proposed modules on the \textit{RIM-ONE-r3} dataset. P-L means vanilla pseudo-labeling method. * represents the accuracy is manually selected from the best epoch. The best results are highlighted.}		\resizebox{0.65\textwidth}{!}{
		\begin{tabular}{cccc|c|c}
			\toprule
			\ P-L\  & \ EMA\  & \ Aug.\  & \ Calib.\ \  & Avg. Dice\ $\uparrow$ & Avg. ASSD\ $\downarrow$ \\
			
			\midrule
			
			\multirow{2}{*}{$\checkmark$} & & & & 64.19 & 15.11 \\
                 & & & & (84.68*) & (9.67*) \\
                 
                 \midrule
                 
                $\checkmark$ & $\checkmark$ & & & 83.63 & 8.51 \\\
                $\checkmark$ & $\checkmark$ & $\checkmark$ & & 84.36 & 8.48 \\
                $\checkmark$ & $\checkmark$ & & $\checkmark$ & 86.04 & 8.26 \\
                $\checkmark$ & $\checkmark$ & $\checkmark$ & $\checkmark$ & \textbf{87.26} & \textbf{7.29} \\
			
			\bottomrule
		\end{tabular}}
	\label{tab2}
	% \vspace{-3mm}
\end{table}

% \vspace{-2mm}
\subsection{Experimental Results}%\vspace{-1mm}
The quantitative evaluation results are shown in Tab.~\ref{tab1}. We include the without adaptation results from \cite{chen2021source} as a lower bound, and the supervised learning results from \cite{wang2019boundary} as an upper bound, same as \cite{chen2021source}. As shown in the table, both two quantitative metric results perform better than previous state-of-the-art SFDA methods and even show an improvement against traditional UDA methods on some metrics. Especially in the RIM-ONE-r3 dataset, our CBMT gains a great performance increase than previous works (dice gains by 3.23 on disc), because the domain shift issue is severer here and has big potential for improvement. 

Moreover, CBMT alleviates the need for precise tuning of hyper-parameters. Here we could set a relatively long training procedure (our epoch number is 10 times that of \cite{chen2021source,xu2022denoising}), and safely select the last checkpoint as our final result without concerning about model degradation issue, which is crucial in real-world clinical source-free domain adaptation application.

% \vspace{-2mm}
\subsection{Further Analyses}%\vspace{-1mm}
\label{Further_Analyses}

\noindent\textbf{Ablation study.} In order to assess the contribution of each component to the final performance, we conducted an ablation study on the main modules of CBMT, as summarized in Table~\ref{tab2}. Note that we reduced the learning rates by a factor of 20 for the experiments of the vanilla pseudo-labeling method to get comparable performance because models are prone to degradation without EMA updating. As observed in quantitative results, the EMA update strategy avoids the model from degradation, which the vanilla pseudo-labeling paradigm suffers from. Image augmentation and loss calibration also boost the model accuracy, and the highest performance is achieved with both. The loss calibration module achieves more improvement in its solution to class imbalance, while image augmentation is easy to implement and plug-and-play under various circumstances.

\begin{figure}[t]
    \centering
    \setlength{\abovecaptionskip}{0cm}
    \begin{subfigure}[b]{0.49\linewidth}
        \centering
        \includegraphics[width=\linewidth]{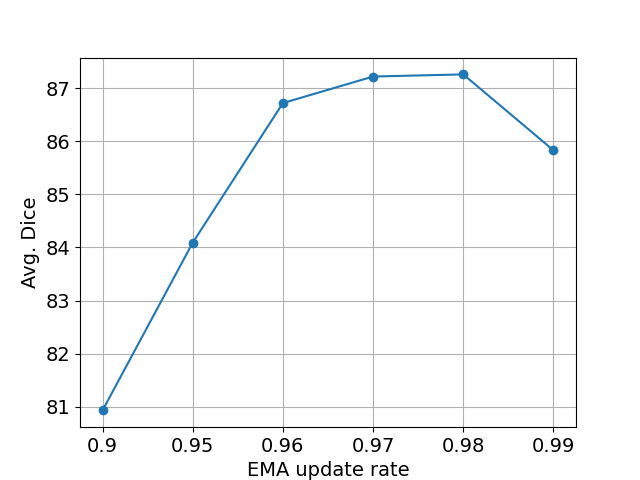}
        \caption{}
        \label{fig2-1}
    \end{subfigure}
    \hfill
    \begin{subfigure}[b]{0.49\linewidth}
        \centering
        \includegraphics[width=\linewidth]{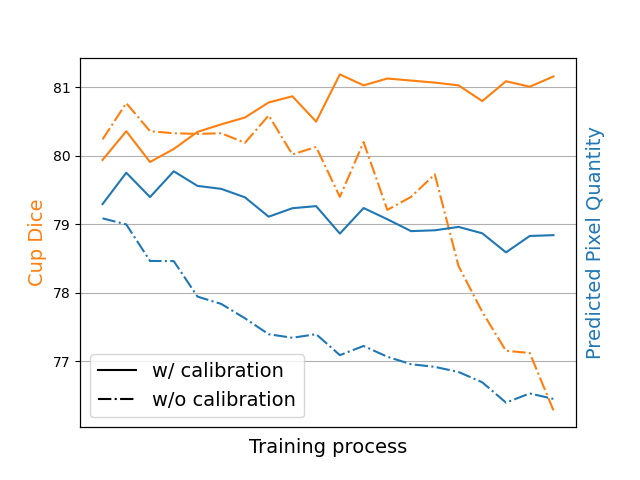}
        \caption{}
        \label{fig2-2}
    \end{subfigure}
    \caption{(a) Model performance with different EMA update rate $\lambda$ setting. (b) Training curves with and without our proposed loss calibration scheme.}
    \label{fig2}
    % \vspace{-5mm}
\end{figure}

\noindent\textbf{Hyper-parameter sensitivity analysis.} We further investigate the impact of different hyper-parameter. Fig~\ref{fig2}(a) presents the accuracy with different EMA update rate parameters $\lambda$. It demonstrates that both too low and too high update rates would cause a drop in performance, which is quite intuitive: a higher $\lambda$ leads to inconsistency between the teacher and student, and thus teacher can hardly learn knowledge from the student; On the other hand, a lower $\lambda$ will always keep teacher and student close, making it degenerated to vanilla pseudo-labeling. But within a reasonable range, the model is not sensitive to update rate $\lambda$.

\begin{table}[t]
	\centering
	\caption{Loss calibration weight with different thresholds $\alpha$ on \textit{RIM-ONE-r3} dataset. Our method is robust to the hyper-parameter setting.}
		\begin{tabular}{c|>{\centering\arraybackslash}p{1cm}>{\centering\arraybackslash}p{1cm}>{\centering\arraybackslash}p{1cm}>{\centering\arraybackslash}p{1cm}>{\centering\arraybackslash}p{1cm}>{\centering\arraybackslash}p{1cm}>{\centering\arraybackslash}p{1cm}>{\centering\arraybackslash}p{1cm}>{\centering\arraybackslash}p{1cm}>{\centering\arraybackslash}p{1cm}}
			\toprule
   
			$\alpha$ & 0 & 0.1 & 0.2 & 0.3 & 0.4 & 0.5 & 0.6 & 0.7 & 0.8 & 0.9 \\
			
			\midrule
   
			$\eta_k^{\text{fg}}/\eta_k^{\text{bg}}$ & 2.99 & 0.24 & 0.24 & 0.24 & 0.24 & 0.24 & 0.23 & 0.23 & 0.22 & 0.22 \\
                
			\bottomrule
		\end{tabular}
	\label{tab3}
	% \vspace{-4mm}
\end{table}

To evaluate the variation of the loss calibration weight $\eta_k^{\text{fg}}/\eta_k^{\text{bg}}$ with different constraint thresholds $\alpha$, we present the results in Tab.~\ref{tab3}. As we discussed in Sec.~\ref{loss_calibration}, most pixels in an image are well-classified, and if we simply calculate with all pixels (i.e. $\alpha=0$), as shown in the first column, the mean loss of background will be severely underestimated due to the large quantity of zero-loss pixel. Besides, as $\alpha$ changes, the calibration weight varies little, indicating the robustness of our calibration technique to threshold $\alpha$.

\noindent\textbf{The effectiveness of loss calibration to balance class.} The class imbalance problem can cause misalignment in the learning processes of different classes, leading to a gradual decrease of predicted foreground area. This can ultimately result in model degradation. As shown in Fig.~\ref{fig2}(b), neglecting the issue of class imbalance can cause a significant drop in the predicted pixel quantity of the class ``cup'' during training, and finally leads to a performance drop. Loss calibration performs a theoretical investigation and proposes an effective technique to alleviate this issue by balancing loss with global context.

% \vspace{-2mm}
\section{Conclusion}
% \vspace{-2mm}
In this work, we propose a class-balanced mean teacher framework to realize robust SFDA learning for more realistic clinical application. Based on the observation that model suffers from degradation issues during adaptation training, we introduce a mean teacher strategy to update the model via an exponential moving average way, which alleviates error accumulation. Meanwhile, by investigating the foreground and background imbalance problem, we present a global knowledge guided loss calibration module. Experiments on two fundus image segmentation datasets show that CBMT outperforms previous SFDA methods.

\subsubsection{Acknowledgement.} This work was partly supported by Shenzhen Key Laboratory of next generation interactive media innovative technology (No: ZDSYS202 \\
10623092001004).

% ---- Bibliography ----
\bibliographystyle{splncs04.bst}
\bibliography{mybibliography.bib}

\begin{thebibliography}{10}
\providecommand{\url}[1]{\texttt{#1}}
\providecommand{\urlprefix}{URL }
\providecommand{\doi}[1]{https://doi.org/#1}

\bibitem{bateson2020source}
Bateson, M., Kervadec, H., Dolz, J., Lombaert, H., Ben~Ayed, I.: Source-relaxed
  domain adaptation for image segmentation. In: International Conference on
  Medical Image Computing and Computer-Assisted Intervention. pp. 490--499.
  Springer (2020)

\bibitem{cai2019towards}
Cai, J., Zhang, Z., Cui, L., Zheng, Y., Yang, L.: Towards cross-modal organ
  translation and segmentation: A cycle-and shape-consistent generative
  adversarial network. Medical image analysis  \textbf{52},  174--184 (2019)

\bibitem{carion2020end}
Carion, N., Massa, F., Synnaeve, G., Usunier, N., Kirillov, A., Zagoruyko, S.:
  End-to-end object detection with transformers. In: European conference on
  computer vision. pp. 213--229. Springer (2020)

\bibitem{chen2021source}
Chen, C., Liu, Q., Jin, Y., Dou, Q., Heng, P.A.: Source-free domain adaptive
  fundus image segmentation with denoised pseudo-labeling. In: International
  Conference on Medical Image Computing and Computer-Assisted Intervention. pp.
  225--235. Springer (2021)

\bibitem{chen2018encoder}
Chen, L.C., Zhu, Y., Papandreou, G., Schroff, F., Adam, H.: Encoder-decoder
  with atrous separable convolution for semantic image segmentation. In:
  Proceedings of the European conference on computer vision (ECCV). pp.
  801--818 (2018)

\bibitem{drozdzal2016importance}
Drozdzal, M., Vorontsov, E., Chartrand, G., Kadoury, S., Pal, C.: The
  importance of skip connections in biomedical image segmentation. In: Deep
  learning and data labeling for medical applications, pp. 179--187. Springer
  (2016)

\bibitem{fumero2011rim}
Fumero, F., Alay{\'o}n, S., Sanchez, J.L., Sigut, J., Gonzalez-Hernandez, M.:
  Rim-one: An open retinal image database for optic nerve evaluation. In: 2011
  24th international symposium on computer-based medical systems (CBMS).
  pp.~1--6. IEEE (2011)

\bibitem{gadermayr2019generative}
Gadermayr, M., Gupta, L., Appel, V., Boor, P., Klinkhammer, B.M., Merhof, D.:
  Generative adversarial networks for facilitating stain-independent supervised
  and unsupervised segmentation: a study on kidney histology. IEEE transactions
  on medical imaging  \textbf{38}(10),  2293--2302 (2019)

\bibitem{ganin2016domain}
Ganin, Y., Ustinova, E., Ajakan, H., Germain, P., Larochelle, H., Laviolette,
  F., Marchand, M., Lempitsky, V.: Domain-adversarial training of neural
  networks. The journal of machine learning research  \textbf{17}(1),
  2096--2030 (2016)

\bibitem{he2023camouflaged}
He, C., Li, K., Zhang, Y., Tang, L., Zhang, Y., Guo, Z., Li, X.: Camouflaged
  object detection with feature decomposition and edge reconstruction. In:
  Proceedings of the IEEE/CVF Conference on Computer Vision and Pattern
  Recognition. pp. 22046--22055 (2023)

\bibitem{he2023weakly}
He, C., Li, K., Zhang, Y., Xu, G., Tang, L., Zhang, Y., Guo, Z., Li, X.:
  Weakly-supervised concealed object segmentation with sam-based pseudo
  labeling and multi-scale feature grouping. arXiv preprint arXiv:2305.11003
  (2023)

\bibitem{ibtehaz2020multiresunet}
Ibtehaz, N., Rahman, M.S.: Multiresunet: Rethinking the u-net architecture for
  multimodal biomedical image segmentation. Neural Networks  \textbf{121},
  74--87 (2020)

\bibitem{javanmardi2018domain}
Javanmardi, M., Tasdizen, T.: Domain adaptation for biomedical image
  segmentation using adversarial training. In: 2018 IEEE 15th International
  Symposium on Biomedical Imaging (ISBI 2018). pp. 554--558. IEEE (2018)

\bibitem{kamnitsas2017unsupervised}
Kamnitsas, K., Baumgartner, C., Ledig, C., Newcombe, V., Simpson, J., Kane, A.,
  Menon, D., Nori, A., Criminisi, A., Rueckert, D., et~al.: Unsupervised domain
  adaptation in brain lesion segmentation with adversarial networks. In:
  International conference on information processing in medical imaging. pp.
  597--609. Springer (2017)

\bibitem{karani2021test}
Karani, N., Erdil, E., Chaitanya, K., Konukoglu, E.: Test-time adaptable neural
  networks for robust medical image segmentation. Medical Image Analysis
  \textbf{68},  101907 (2021)

\bibitem{li2020adversarial}
Li, K., Zhang, Y., Li, K., Fu, Y.: Adversarial feature hallucination networks
  for few-shot learning. In: Proceedings of the IEEE/CVF conference on computer
  vision and pattern recognition. pp. 13470--13479 (2020)

\bibitem{liu2021adapting}
Liu, X., Xing, F., Yang, C., El~Fakhri, G., Woo, J.: Adapting off-the-shelf
  source segmenter for target medical image segmentation. In: International
  Conference on Medical Image Computing and Computer-Assisted Intervention. pp.
  549--559. Springer (2021)

\bibitem{liu2021unbiased}
Liu, Y.C., Ma, C.Y., He, Z., Kuo, C.W., Chen, K., Zhang, P., Wu, B., Kira, Z.,
  Vajda, P.: Unbiased teacher for semi-supervised object detection. arXiv
  preprint arXiv:2102.09480  (2021)

\bibitem{milletari2016v}
Milletari, F., Navab, N., Ahmadi, S.A.: V-net: Fully convolutional neural
  networks for volumetric medical image segmentation. In: 2016 fourth
  international conference on 3D vision (3DV). pp. 565--571. IEEE (2016)

\bibitem{orlando2020refuge}
Orlando, J.I., Fu, H., Breda, J.B., van Keer, K., Bathula, D.R., Diaz-Pinto,
  A., Fang, R., Heng, P.A., Kim, J., Lee, J., et~al.: Refuge challenge: A
  unified framework for evaluating automated methods for glaucoma assessment
  from fundus photographs. Medical image analysis  \textbf{59},  101570 (2020)

\bibitem{ren2015faster}
Ren, S., He, K., Girshick, R., Sun, J.: Faster r-cnn: Towards real-time object
  detection with region proposal networks. Advances in neural information
  processing systems  \textbf{28} (2015)

\bibitem{ronneberger2015u}
Ronneberger, O., Fischer, P., Brox, T.: U-net: Convolutional networks for
  biomedical image segmentation. In: International Conference on Medical image
  computing and computer-assisted intervention. pp. 234--241. Springer (2015)

\bibitem{sandler2018mobilenetv2}
Sandler, M., Howard, A., Zhu, M., Zhmoginov, A., Chen, L.C.: Mobilenetv2:
  Inverted residuals and linear bottlenecks. In: Proceedings of the IEEE
  conference on computer vision and pattern recognition. pp. 4510--4520 (2018)

\bibitem{sivaswamy2015comprehensive}
Sivaswamy, J., Krishnadas, S., Chakravarty, A., Joshi, G., Tabish, A.S.,
  et~al.: A comprehensive retinal image dataset for the assessment of glaucoma
  from the optic nerve head analysis. JSM Biomedical Imaging Data Papers
  \textbf{2}(1), ~1004 (2015)

\bibitem{vu2019advent}
Vu, T.H., Jain, H., Bucher, M., Cord, M., P{\'e}rez, P.: Advent: Adversarial
  entropy minimization for domain adaptation in semantic segmentation. In:
  Proceedings of the IEEE/CVF Conference on Computer Vision and Pattern
  Recognition. pp. 2517--2526 (2019)

\bibitem{wang2019boundary}
Wang, S., Yu, L., Li, K., Yang, X., Fu, C.W., Heng, P.A.: Boundary and
  entropy-driven adversarial learning for fundus image segmentation. In:
  International Conference on Medical Image Computing and Computer-Assisted
  Intervention. pp. 102--110. Springer (2019)

\bibitem{xu2022denoising}
Xu, Z., Lu, D., Wang, Y., Luo, J., Wei, D., Zheng, Y., Tong, R.K.y.: Denoising
  for relaxing: Unsupervised domain adaptive fundus image segmentation without
  source data. In: Medical Image Computing and Computer Assisted
  Intervention--MICCAI 2022: 25th International Conference, Singapore,
  September 18--22, 2022, Proceedings, Part V. pp. 214--224. Springer (2022)

\end{thebibliography}


\begin{thebibliography}{1}
\providecommand{\url}[1]{\texttt{#1}}
\providecommand{\urlprefix}{URL }
\providecommand{\doi}[1]{https://doi.org/#1}

\bibitem{xu2022denoising}
Xu, Z., Lu, D., Wang, Y., Luo, J., Wei, D., Zheng, Y., Tong, R.K.y.: Denoising
  for relaxing: Unsupervised domain adaptive fundus image segmentation without
  source data. In: Medical Image Computing and Computer Assisted
  Intervention--MICCAI 2022: 25th International Conference, Singapore,
  September 18--22, 2022, Proceedings, Part V. pp. 214--224. Springer (2022)

\end{thebibliography}

\end{document}

% \section{Supplementary Material}

%\subsection{Experimental Results}

% \vspace{-1cm}
\begin{table}[t]
	\centering
	\caption{Quantitative results of comparison with different methods (source domain: Drishti-GS dataset; target domain: RIM-ONE-r3 dataset and REFUGE dataset, following the setting in \cite{xu2022denoising}), and the best scores of each columns is highlighted. $\pm$ refers to the standard deviation across samples in the dataset. S-F means source-free.}
	\vspace{-0.5cm}
	\label{tab1}
	\begin{center}
		\setlength\tabcolsep{3.0pt}
		\resizebox{0.85\textwidth}{!}{%
			\begin{tabular}{l|c|c|c|c|c}
				\toprule[1.5pt]
				
				\multirow{2}{*}{Methods} & \multirow{2}{*}{S-F} & \multicolumn{2}{c|}{Optic Disc Segmentation} & \multicolumn{2}{c}{Optic Cup Segmentation} \\
				\cline{3-6}
				{} & {} & Dice[\%]\ $\uparrow$ & ASSD[pixel]\ $\downarrow$ & Dice[\%]\ $\uparrow$ & ASSD[pixel]\ $\downarrow$ \\
				\cline{1-6}
				\multicolumn{6}{l}{\emph{RIM-ONE-r3}}\\
				\hline
				W/o DA & & 89.15$\pm${14.07} & 9.62$\pm${9.95} & 57.33$\pm${35.36} & 10.92$\pm${6.99} \\
				Upper bound & & 95.88$\pm${2.21} & 3.54$\pm${1.82} & 78.34$\pm${20.88} & 7.51$\pm${5.89} \\
				\hline
				\hline
				BEAL & $\times$ & 90.69$\pm${6.93} & 8.90$\pm${7.69} & 70.47$\pm${27.40} & 12.68$\pm${14.24} \\
				AdvEnt & $\times$ & 91.88$\pm${9.87} & 6.97$\pm${5.27} & 71.23$\pm${26.61} & \textbf{10.40$\pm${10.79}} \\
				SRDA & $\checkmark$ & 90.62$\pm${15.48} & 8.85$\pm${7.93} & 62.08$\pm${21.48} & 15.83$\pm${8.51} \\
				TENT & $\checkmark$ & 90.25$\pm${10.41} & 8.28$\pm${6.95} & 62.89$\pm${19.83} & 14.83$\pm${8.62} \\
				DPL & $\checkmark$ & 91.48$\pm${7.09} & 8.95$\pm${7.13} & 64.01$\pm${19.88} & 16.39$\pm${7.53} \\
                U-D4R & $\checkmark$ & 93.31$\pm${5.41} & 6.63$\pm${6.05} & 68.59$\pm${13.87} & 11.62$\pm${5.66} \\
				\hline
				\textbf{CBMT(Ours)} & $\checkmark$ & \textbf{94.37$\pm${2.84}} & \textbf{5.04$\pm${2.61}} & \textbf{71.93$\pm${18.64}} & 11.98$\pm${7.11} \\
				\hline

				\multicolumn{6}{l}{\emph{REFUGE}}\\
				\hline
				W/o DA & & 89.68$\pm${5.62} & 9.03$\pm${4.04} & 79.45$\pm${10.74} & 8.95$\pm${4.68} \\
				Upper bound & & 94.79$\pm${2.01} & 4.70$\pm${1.55} & 87.67$\pm${6.60} & 5.78$\pm${3.67} \\
				\hline
				\hline
				BEAL & $\times$ & 91.43$\pm${3.79} & 7.93$\pm${3.85} & 83.24$\pm${10.85} & 8.24$\pm${6.86} \\
				AdvEnt & $\times$ & 90.39$\pm${3.92} & 8.68$\pm${3.99} & 83.66$\pm${11.03} & 8.07$\pm${7.08} \\
				SRDA & $\checkmark$ & 86.41$\pm${2.93} & 9.56$\pm${4.13} & 81.57$\pm${7.63} & 8.94$\pm${4.82} \\
				TENT & $\checkmark$ & 92.04$\pm${2.37} & 8.25$\pm${4.58} & 83.58$\pm${8.42} & 8.01$\pm${4.26} \\
				DPL & $\checkmark$ & 92.57$\pm${2.81} & 7.83$\pm${3.92} & 84.02$\pm${7.84} & 7.56$\pm${4.05} \\
                U-D4R & $\checkmark$ & 93.60$\pm${2.41} & 7.09$\pm${4.59} & \textbf{85.11$\pm${7.77}} & \textbf{6.83$\pm${4.64}} \\
				\hline
				\textbf{CBMT(Ours)} & $\checkmark$ & \textbf{94.21$\pm${4.29}} & \textbf{5.73$\pm${4.31}} & 84.10$\pm${8.29} & 8.38$\pm${6.52} \\
				% \hline
				
				\bottomrule[1.5pt]
		    \end{tabular}}
	\end{center}
	\vspace{-1.5cm}
\end{table}

\begin{figure}[h]
% \vspace{-0.8cm}
\begin{center}
    \includegraphics[width=0.6\textwidth]{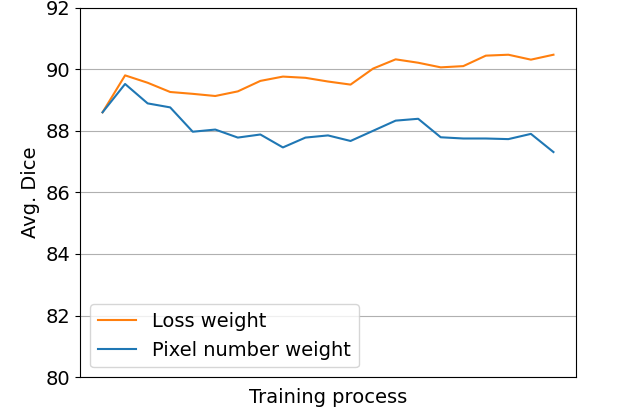}
    \vspace{-0.4cm}
    \caption{Comparison of different loss calibration implementations on Drishti-GS dataset. Loss weight refers to the method mentioned in our paper, and pixel number weight means only utilizing pixel number of foreground and background to compute $\eta_k^{\text{fg}}$ and $\eta_k^{\text{bf}}$. Our method leverages more informative pixels and achieves higher performance.} \vspace{-0.4cm}
\label{fig_loss}
\end{center}
\end{figure}

\clearpage

\begin{table}[h]
    \centering
    \caption{Strong-weak augmentation ablation study results on RIM-ONE-r3 dataset. Teacher model generates more accurate pseudo labels with weakly-augmented images, while student model learns more consistent feature representation with strong ones.}
    \vspace{-0.2cm}
    % \resizebox{0.65\textwidth}{!}{
		\begin{tabular}{>{\centering\arraybackslash}p{3cm}|>{\centering\arraybackslash}p{3cm}|>{\centering\arraybackslash}p{3cm}}
			\toprule
			Teacher input & Student input & Avg. Dice \\
			
			\midrule
			
			\multirow{2}{*}{Weak-aug} & Weak-aug & 86.04 \\
                % \cline{2-3}
                 & Strong-aug & \textbf{87.26} \\
                 
                 \midrule
                 
                \multirow{2}{*}{Strong-aug} & Weak-aug & 86.16 \\
                % \cline{2-3}
                 & Strong-aug & 86.33 \\
			
			\bottomrule
		\end{tabular}
    % }
	\label{tab2}
	\vspace{-6mm}
\end{table}

\begin{figure}[h]
% \vspace{-0.8cm}
\begin{center}
    \includegraphics[width=0.8\textwidth]{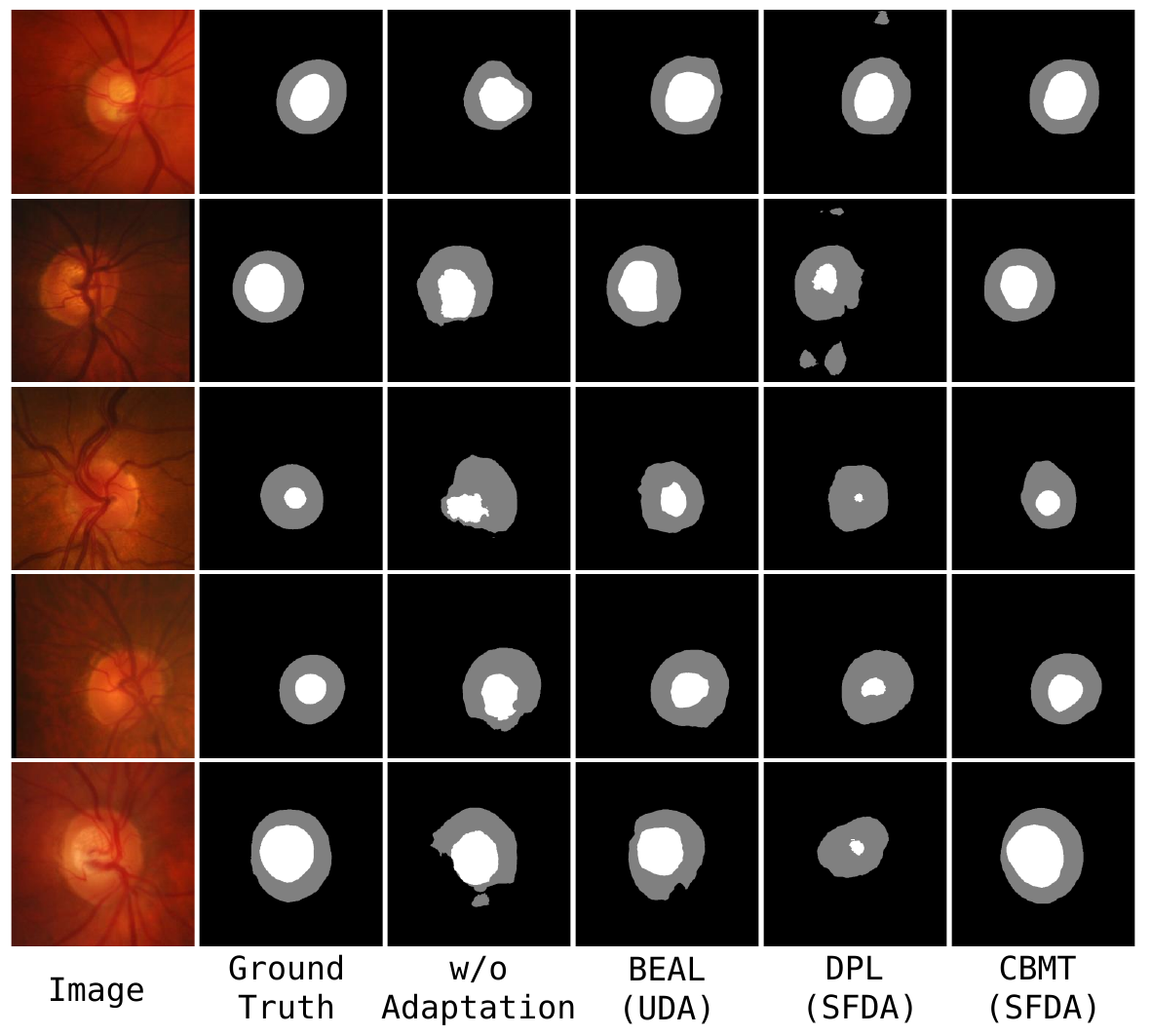}
    \vspace{-0.2cm}
    \caption{Visualizations of adaptation performance with different methods.} 
\label{fig_visual}
\end{center}
\end{figure}

%\subsection{Strong-weak Augmentation Ablation Study}

% \clearpage

%\subsection{Loss Calibration Implementation}

%\subsection{Visual Comparison}

\bibliographystyle{splncs04.bst}
\bibliography{mybibliography.bib}